\documentclass[letterpaper, 10 pt, conference]{ieeeconf}  

\IEEEoverridecommandlockouts                              

\usepackage{graphicx}
\usepackage{amssymb}
\usepackage{amsmath}
\usepackage{bm}
\usepackage{xspace}
\usepackage{makecell}
\usepackage{booktabs}
\usepackage{url}
\usepackage{subfigure}
\usepackage{tabularx}
\usepackage{multirow}
\usepackage{multicol}
\usepackage{threeparttable}
\usepackage{placeins}
\usepackage{balance}

\usepackage{cite}
\usepackage{caption}
\makeatletter
\let\NAT@parse\undefined
\makeatother
\usepackage{hyperref}

\DeclareMathOperator*{\argmin}{arg\,min}

\newcommand{\ie}{i.e.,\ }
\newcommand{\eg}{e.g.,\ }

\newcommand{\etal}{\xspace{}et al.\xspace}

\usepackage{color}
\usepackage{xcolor}

\newcommand{\sub}[2]{{#2}}

\title{\LARGE \bf
World Model-based Perception for Visual Legged Locomotion
}

\author{Hang Lai$^{1}$$^\dagger$, \hspace{-5.0pt} Jiahang Cao$^{1}$,\hspace{-0.0pt} Jiafeng Xu$^{2}$$^{\ast}$,\hspace{-0.0pt} Hongtao Wu$^{2}$,\hspace{-1.0pt} Yunfeng Lin$^{1}$$^\dagger$,\hspace{-1.0pt} Tao Kong$^{2}$,\hspace{-1.0pt} Yong Yu$^{1}$,\hspace{-1.0pt} Weinan Zhang$^{1}$$^{\ast}$
\thanks{$^\dagger$Work done during internship at Bytedance Research.}
\thanks{$^\ast$Corresponding author.}
\thanks{$^{1}$Dept. of Computer Sci. and Eng., Shanghai Jiao Tong University, China.}
\thanks{$^{2}$ByteDance Research, China.}%
}

\begin{document}
	
\maketitle
\thispagestyle{empty}
\pagestyle{empty}

\begin{abstract}
Legged locomotion over various terrains is challenging and requires precise perception of the robot and its surroundings from both proprioception and vision. However, learning directly from high-dimensional visual input is often data-inefficient and intricate. To address this issue, traditional methods attempt to learn a teacher policy with access to privileged information first and then learn a student policy to imitate the teacher's behavior with visual input. Despite some progress, this imitation framework prevents the student policy from achieving optimal performance due to the information gap between inputs. Furthermore, the learning process is unnatural since animals intuitively learn to traverse different terrains based on their understanding of the world without privileged knowledge. Inspired by this natural ability, we propose a simple yet effective method, World Model-based Perception (WMP), which builds a world model of the environment and learns a policy based on the world model. We illustrate that though completely trained in simulation, the world model can make accurate predictions of real-world trajectories, thus providing informative signals for the policy controller. Extensive simulated and real-world experiments demonstrate that WMP outperforms state-of-the-art baselines in traversability and robustness. Videos and Code are available at: \href{https://wmp-loco.github.io/}{https://wmp-loco.github.io/}.


\end{abstract}

\section{Introduction}
\label{sec:Introduction}
Reinforcement Learning (RL) has recently achieved remarkable success in legged locomotion across diverse terrains by training a policy in physical simulation and then transferring it to the real world (\ie sim-to-real transfer) \cite{tan2018sim, yu2019sim}. Typically, such RL policy takes the proprioception (\eg positions and velocities of joints) or visual image as input and outputs the desired position or effort control for each actuated joint \cite{imitate, hwangbo2019learning}. Previous literature has shown that a blind policy with only proprioceptive input can traverse terrains like slopes and stairs \cite{rma, tert, dreamwaq, amp-wu, him}, but fails in more challenging ones like gaps or pits \cite{egocentric, parkour2, parkour1}, where a robot must perceive such terrain in advance; therefore, visual image perception is indispensable \cite{wild, vision-bipedal}.

However, directly learning a policy with dense pixel input using reward signals is extremely data-inefficient \cite{anand2019unsupervised}. Moreover, with a forward-facing camera, a policy needs to remember past perceptions to anticipate the terrain under the robot's feet \cite{egocentric}, which poses an additional challenge for policy learning. To facilitate policy training, the \emph{privileged learning} framework \cite{terrain} is proposed, which decomposes the training process into two phases. First, a teacher policy is trained with access to low-dimensional privileged information like the scandots around the robot, which is usually inaccessible in the real world. Afterward, a student policy is trained to mimic the teacher's actions based on the seen images via ConvNet-RNN architecture \cite{egocentric, parkour1, parkour2}.

\begin{figure}[tb]
	\centering
        \vspace{-2pt}
         \hspace{-25pt}
	\includegraphics[width=0.53\textwidth]{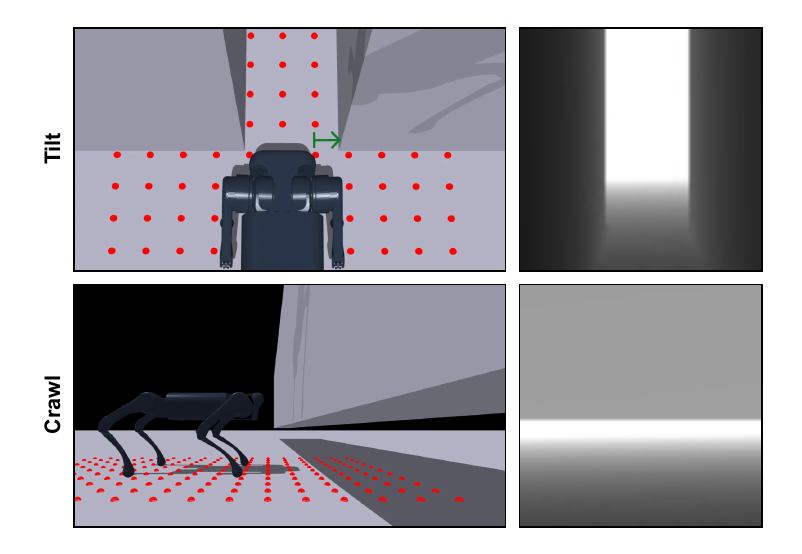}
	\vspace{-10pt}
        \hspace{-20pt}
	\caption{Scandots (Left) and the corresponding depth images (Right). Top: Sparse scandots\protect\footnotemark can not distinguish the precise distance to the boundaries (indicated in green), leading to a collision with the left barrier. Bottom: Scandots can not represent off-ground obstacles. In contrast, depth images can represent these terrains well.}
	
	\vspace{-15pt}
	\label{fig:scandots}
\end{figure}

\footnotetext{\sub{Due to the limitations of the simulator, we cannot construct scandots denser }{The scandots are no denser} than the terrain mesh vertices, which have a horizontal resolution of 10cm in our simulation setting.}

Privileged learning has been widely exploited and exhibits traversability over multiple terrains in quadrupedal locomotion \cite{rma, egocentric, wild, parkour2}, but it still has some limitations. Firstly, since the student cannot recover the teacher's behavior perfectly due to generalization error \cite{mohri2018foundations}, the performance of student policy tends to lag behind the teacher's. The performance discrepancy is further magnified when an information gap exists between privileged information and images \cite{hib}. Secondly, the teacher policy needs to access different types of extra information, which is labor-intensive to design. For example, Cheng \etal~\cite{parkour2} choose the terrain scandots as the privileged information, which can not be generalized to terrains that require precise boundary distinction, like Tilt, or ones with off-ground obstacles, like Crawl, as illustrated in Figure~\ref{fig:scandots}. Besides, Zhuang \etal~\cite{parkour1} use the geometry of obstacles as privileged information, which is terrain-relevant. Therefore, they train different teachers separately for each terrain. These limitations may hinder its broader applications in more complex scenarios.

In contrast, animals naturally learn to traverse unstructured fields and can make good decisions in unfamiliar situations with limited perception. One hypothesis is that animals, especially humans, build a mental model that holds their understanding of the real world~\cite{forrester1971counterintuitive, ha2018world}. When performing actions, it helps perceive past information and predict future sensory data \cite{keller2012sensorimotor, maus2013motion, nortmann2015primary, leinweber2017sensorimotor}. Inspired by these findings, model-based RL (MBRL) strives to learn a world model from collected data and derives a policy from it \cite{luo2024survey}. With the help of such models, MBRL has made immense progress in a variety of tasks with limited data, ranging from simulated robot control \cite{dreamerv1, mbpo} to playing video games \cite{ha2018world, dreamerv2, dreamerv3}. However, the application of world models in vision-based legged locomotion is still lacking.

This paper investigates whether visual legged locomotion can benefit from world model learning. To this end, we present \textbf{W}orld \textbf{M}odel-based \textbf{P}erception (WMP), a novel end-to-end framework combining advanced MBRL with sim-to-real transfer. Specifically, WMP trains a world model in simulations to predict future perceptions using past ones and learns a policy given the abstract description extracted from the world model. Though trained entirely using simulated data, the world model can still predict real-world perception well. By leveraging the learned model, WMP circumvents the limitations of privileged learning and naturally compresses a series of high-dimensional perceptions into a meaningful representation, contributing to decision-making.

We compare WMP to state-of-the-art baselines over multiple terrains, including terrains like Slope and Stair and more difficult ones like Gap and Crawl. In simulation comparison, WMP obtains near-optimal rewards compared with the teacher policy, surpassing the student policy by a pronounced margin. Subsequently, we evaluate WMP and baselines on a real Unitree A1 robot, where WMP successfully traverses the tested terrains with increased difficulty, verifying the advantage of world model learning in robot control. For example, WMP can traverse Gap with 85cm (about 2.1x robot length), Climb with 55cm (about 2.2x robot height), and Crawl with 22cm (about 0.8x robot height), achieving the best traversal performance on the A1 robot. To the best of our knowledge, this is the first work that deals with challenging vision-based legged locomotion via world modeling, which could become a new paradigm for robot control tasks.

\section{Related Work}
\label{sec:Related_Work}

\noindent \textbf{RL for Legged Locomotion.}
Reinforcement learning (RL) has emerged as a promising method for legged locomotion~\cite{tan2018sim,yu2019sim,hwangbo2019learning,imitate,minimizing}. Previous literature has shown that policies with only proprioception as input can go through multiple terrains in the real world by leveraging biologically inspired rewards design\cite{terrain, rma, hwangbo2019learning}, domain randomization \cite{imitate, randomization2, dreamwaq}, and curriculum learning \cite{leggedgym, amp-wu}. However, without visual perception, it can be extremely difficult for these "blind" robots to tackle more complex terrains \cite{wild, parkour1}. To incorporate visual information, the privileged learning framework is developed \cite{egocentric, parkour1, parkour2}, where a teacher policy trained with access to privileged information is used to guide a vision-based student policy. However, the design of privileged information and the performance gap between teacher and student remain critical limitations of these methods \cite{pinto2017asymmetric, tert}. Besides, one concurrent work proposes to estimate the scandots using past observations \cite{pie}, which also suffers from the limitation of scandots inevitably. In contrast, Our method presents a more general and effective framework to incorporate visual images without these limitations.

\vspace{0.1cm}
\noindent \textbf{Model-based RL.} Model-based reinforcement learning (MBRL) approaches have shown promise for learning complex robot control policies by learning a dynamics model of the environment to help decision-making \cite{pets, slbo, mbpo, luo2024survey}. In MBRL, much effort has been devoted to learning an accurate model in partially observable and pixel-input environments \cite{ha2018world, RSSM, dreamerv1, dreamerv2}. As a prominent example, Dreamer-V3 \cite{dreamerv3} achieves impressive performance across diverse domains by learning a world model in a compact latent space, namely the Recurrent State-Space Model (RSSM). Inspired by the success of Dreamer, RSSM has also been widely exploited in robot control tasks, ranging from robotic manipulation \cite{Yamada2023twist, ferraro2023focus, Mendonca2023Structed} to blind quadrupedal locomotion \cite{Daydreamer}. Our work also builds upon the RSSM world model architecture, which can seamlessly take advantage of innovations in MBRL literature and push the boundaries of its application in legged locomotion with visual input.

\section{Preliminaries}
\label{sec:Preliminaries}

\begin{figure*}[th!]
\vspace{6pt}
\centering
\includegraphics[width=0.93\textwidth]{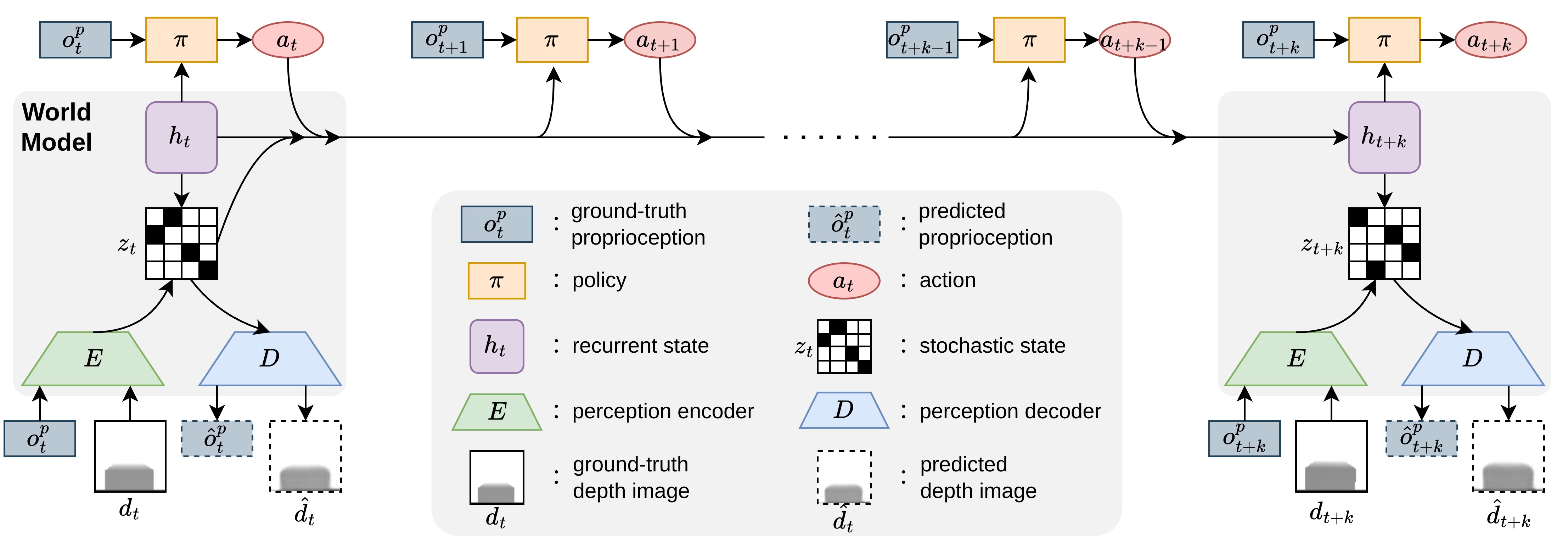}
\vspace{-5pt}	
 \caption{Illustration of the WMP framework. The world model runs at a lower frequency than policy, with an update interval of $k$ timesteps.}
	\vspace{-10pt}
	\label{fig:framework}
\end{figure*}

We formulate legged locomotion as a Partially Observable Markov Decision Process (POMDP) defined by the tuple ($\mathcal{S}, \mathcal{O}, \mathcal{A}, T, r, \gamma$), where $\mathcal{S}$, $\mathcal{O}$, and $\mathcal{A}$ are the state, observation, and action spaces, respectively. $T(s_{t+1}\mid s_t,a_t)$ is the transition density of state $s_t$ given action $a_t$, and the reward function is denoted as $r(s_t,a_t)$. $\gamma \in (0,1)$ is a discount factor. At each timestep $t$, only the partial observation $o_t \in \mathcal{O}$ can be observed instead of $s_t$ due to the limitation of sensors. The goal of reinforcement learning (RL) is to find the optimal policy $\pi^* \colon \mathcal{O} \to \mathcal{A}$ that maximizes the expected return (sum of discounted rewards):
\begin{equation}
\label{eq: rl-obj}
\pi^* \!\colon\!\!\!\!= \mathop{\arg \max}_\pi \mathbb{E}_{s_{t+1} \sim T(\cdot \mid s_t,a_t), a_t \sim \pi(\cdot \mid o_t)} \Big[\sum_{t=0}^\infty \gamma^t r(s_t,a_t) \Big].
\end{equation}
Specifically, in the vision-based legged locomotion, $o_{t}$ consists of the proprioception $o^{p}_{t}$ and the depth image $d_t$. 
In addition to $o_{t}$, the underlying state $s_{t}$ also contains the privileged information $s^{\mathrm{pri}}_{t}$, which can only be accessed in the simulator:
\begin{align}
o_{t} \colon\!\!\!&= (o^{p}_{t}, d_t), 
\\ s_{t} \colon\!\!\!&= (o_{t}, s^{\mathrm{pri}}_{t}).
\end{align}

\section{Method}
\label{sec:Method}

This section introduces World Model-based Perception (WMP), an end-to-end framework that utilizes world models to extract information from high-dimensional sensor input.
Unlike the two-stage training process used in privileged learning, WMP only adopts one stage to learn the world model and policy simultaneously.

\subsection{World Model Learning}

Following previous works \cite{RSSM, dreamerv3}, we adopt a Recurrent State-Space Model (RSSM) variant as our world model architecture. Considering the computational cost of acquiring depth images in the simulator and the time cost of running RSSM on board, we modify the original RSSM by running the world model with a lower frequency than the policy. As Figure~\ref{fig:framework} shows, the world model updates the recurrent state $h_t$ every $k$ timesteps. Formally, the RSSM in our method consists of four components parameterized by $\phi$:
\begin{equation}
    \begin{aligned}
        &\text{Recurrent model:} & \  h_t &\ = \ f_\phi(h_{t-k},z_{t-k},a_{t-k:t-1}) \\
        &\text{Encoder:}& \  z_t &\ \sim \ q_\phi(\cdot \mid h_{t}, o_{t}) \\
        &\text{Dynamic predictor:}  & \ \hat{z}_t &\ \sim \ p_\phi(\cdot \mid h_t) \\
        &\text{Decoder:} & \ \hat{o}_t &\ \sim \ p_\phi(\cdot \mid h_t, z_t).
    \\
    \end{aligned}
\end{equation}
To be more specific, the recurrent model $f_{\phi}$ operates in the low-dimensional latent space $h$ and predicts the deterministic recurrent state $h_t$ based on the previous $h_{t-k}$, sequence of action $a_{t-k:t-1}$, and the previous stochastic state $z_{t-k}$. From the recurrent state $h_t$, the RSSM computes two distributions over stochastic states $z_t$. The posterior state $z_t$ incorporates information from observation $o_t$ through the encoder $q_{\phi}$. The prior state $\hat{z}_t$ aims to predict the posterior without access to $o_t$, enabling the model to anticipate future dynamics without ground-truth observation. The decoder generates the estimated observation $\hat{o}_t$, making it possible to reconstruct the high-dimensional observation. The recurrent model is implemented using the GRU (Gated Recurrent Unit) \cite{GRU} network, and the encoder and decoder utilize the convolutional neural network (CNN) structure for depth image $d_t$ and multi-layer perceptions (MLP) for proprioception observations $o^p_t$. 

Similar to Hafner \etal~\cite{dreamerv3}, we optimize these four ingredients jointly by minimizing the loss over trajectories of length $L$:
\vspace{-3pt}
\begin{equation}
\begin{aligned}
    \mathcal{L}(\phi)\doteq 
\mathbb{E}_{q_\phi}\Big[\sum_{t=n \cdot k}^L
    &-\ln p_\phi(o_t|z_t,h_t) 
    \\ &+ \beta \mathrm{KL}\big[q_\phi(\cdot \mid h_t,o_t) ||p_\phi(\cdot \mid h_t)\big]
\Big],
\end{aligned}
\label{eq:loss-wm}
\end{equation}
where $n$ is a non-negative integer, and $\beta$ is a hyperparameter. The first term in Eq.~(\ref{eq:loss-wm}) is the reconstruction loss, which encourages the posterior $z_t$ to contain sufficient information about $o_t$, while the second $\mathrm{KL}$ term regularizes the prior and posterior to approximate each other, allowing open-loop prediction of future observations based on current $h_t$ and future actions. Please refer to the original Dreamer literature \cite{dreamerv1, dreamerv2, dreamerv3} for more details about RSSM training. 

\begin{table*}[ht]
\vspace{5pt}
\setlength{\tabcolsep}{4pt}
\renewcommand{\arraystretch}{1.05}
\caption{
Comparison results on different terrains in simulation. Results are averaged over 100 trajectories with different difficulties. Bold numbers indicate the best scores among algorithms excluding Teacher. N/A means that the method is not applicable to the terrain.}
\vspace{-3pt}
\centering
\small
\begin{tabular}{c|c|ccccccc}
\toprule
\multicolumn{1}{c}{} & \multicolumn{1}{c}{} & \multicolumn{1}{c}{Slope \!(0-$36^{\circ}$)} & \multicolumn{1}{c}{Stair \!(5-21cm)} & \multicolumn{1}{c}{Gap \!(0-90cm)} & \multicolumn{1}{c}{Climb \!(0-54cm)} & \multicolumn{1}{c}{Tilt \!(32-28cm)} & \multicolumn{1}{c}{Crawl \!(35-21cm)} \\
\midrule
\multirow{5}{*}{Return $\uparrow$} & \bf WMP (ours) &  $\bf 36.55 \pm 0.82$ &  $\bf 35.06 \pm 3.54$ &  $\bf 32.37 \pm 8.24$ &  $34.64 \pm 2.99$ & $\bf 34.73 \pm 3.13$ & $\bf 36.60 \pm 0.33$ \\
 & \bf Teacher     &  $36.70 \pm 1.79$ &  $35.07 \pm 0.68$ &  $32.80 \pm 7.06$ &  $35.32 \pm 3.33$ & N/A & N/A \\
 & \bf Student  &  $36.31 \pm 1.35$ &  $34.93 \pm 1.32$ &  $27.07 \pm 12.38$ &  $33.42 \pm 5.66$ & N/A & N/A \\
 & \bf Blind  &  $33.96 \pm 2.58$ &  $23.56 \pm 12.98$ &  $9.80 \pm 9.51$ &  $14.11 \pm 13.27$ & $3.94 \pm 2.28$ & $8.92 \pm 7.65$ \\
 & \bf WMP w/o Prop &  $36.07 \pm 0.27$ &  $34.62 \pm 3.51$ &  $30.73 \pm 9.93$ &  $\bf 34.67 \pm 1.95$ & $30.78 \pm 7.22$ & $36.41 \pm 1.76$ \\

\midrule
\multirow{5}{*}{\makecell[r]{Tracking \\ Error $\downarrow$}} & \bf WMP (ours) &  $\bf 0.008 \pm 0.006$ &  $\bf 0.013 \pm 0.010$ &  $\bf 0.26 \pm 0.13$ &  $\bf 0.13 \pm 0.08$ & $\bf 0.02 \pm 0.01$ & $\bf 0.006 \pm 0.002$ \\
 & \bf Teacher  &  $0.007 \pm 0.010$ &  $0.012 \pm 0.005$ &  $0.23 \pm 0.21$ &  $0.09 \pm 0.07$ & N/A & N/A \\
 & \bf Student  &  $\bf 0.008 \pm 0.004$ &  $0.015 \pm 0.006$ &  $0.35 \pm 0.28$ &  $0.16 \pm 0.07$ & N/A & N/A \\
 & \bf Blind &  $0.019 \pm 0.011$ &  $0.033 \pm 0.021$ &  $0.33 \pm 0.19$ &  $0.26 \pm 0.07$ & $0.08 \pm 0.06$ & $0.051 \pm 0.022$ \\
 & \bf WMP w/o Prop &  $0.009 \pm 0.003$ &  $0.017 \pm 0.027$ &  $0.28 \pm 0.18$ &  $0.14 \pm 0.09$ & $0.08 \pm 0.10$ & $0.013 \pm 0.006$ \\

\bottomrule
\end{tabular}
\vspace{-15pt}
\label{table:simulation comparison}
\end{table*}

\subsection{Policy Learning}
Policy learning for vision-based locomotion is non-trivial due to the partial observability, as described in Section~\ref{sec:Preliminaries}. However, the recurrent state $h_{t}$ in a well-trained world model encapsulates sufficient information for future prediction, akin to the underlying Markovian state $s_t$. Building on this insight, we train a policy that incorporates $h_{t}$ as input:
\begin{equation}
    {a}_{t+i} \sim {\pi_{\theta}}\big(\cdot \mid o^p_{t+i}, \mathrm{sg}(h_t)\big), \ \forall i \in [0,k-1],
\end{equation}
where $\mathrm{sg}(\cdot)$ represents the stop-gradient operator. We employ the asymmetric actor-critic framework \cite{pinto2017asymmetric}, where the critic can access the privileged information $s^{\mathrm{pri}}_t$:
\begin{equation}
    {v}(s_{t+i}) \sim {V_{\theta}}\big(\cdot \mid o^p_{t+i}, \mathrm{sg}(h_t), s^{\mathrm{pri}}_{t+i}\big), \ \forall i \in [0,k-1].
\end{equation}
We find that the recurrent state $h_t$ plays an essential role in critic learning since the scandots in $s^{\mathrm{pri}}_t$ can not represent some types of terrains like Tilt and Crawl, as discussed in Section~\ref{sec:Introduction}. 

The actor and critic are trained using the data collected in the simulator via the PPO (Proximal Policy Optimization) algorithm \cite{ppo}. Note that we do not utilize the world model to generate rollout data for policy training as in previous MBRL methods \cite{mbpo, Daydreamer}. Because the world model is trained using simulated data, which means it can not generate more accurate data than the simulator, and sampling data in the simulator is efficient enough. Training world models with real-world data is a possible way to make models more accurate, which we leave as future work.

\subsection{Training Details}
\label{sec:Training Details}
\noindent \textbf{Environment.} We implement our method and baselines based upon the \texttt{legged\_gym} codebase \cite{leggedgym}, which leverages the Isaac Gym simulator \cite{isaacgym} to support simulation of thousands of robots in parallel. Specifically, We create 4096 Unitree A1 \cite{unitree} instances on six types of terrains, including Slope, Stair, Gap, Climb, Crawl, and Tilt, each with varying difficulty levels, as listed in Table~\ref{table:simulation comparison}. We adopt the same terrain curriculum as in Rudin \etal~\cite{leggedgym}. All robots are initialized to different terrains with the lowest difficulty in a certain proportion. The robot is moved to a higher level of difficulty once it passes the borders of its terrain or assigned to a lower level if it moves by less than half of the distance required by its target velocity. Robots take actions at a frequency of 50 Hz, \ie 0.02s per timestep. Depth images are computed every $k$ timesteps and sent to the policy with 100ms latency to facilitate sim-to-real transfer. We also randomize the physical parameters to further improve the policy's robustness as in Cheng \etal~\cite{parkour2}.

\vspace{0.1cm}
\noindent \textbf{State and Action Space.} Precisely, the proprioception observation $o^p_t \in \mathbb{R}^{45}$ consists of base angular velocities, gravity projection, commands, positions and velocities of joints, and last action $a_{t-1}$.
The privileged information $s^{\mathrm{pri}}_t$ contains scandots, foot contact forces, and randomized physical parameters. $d_t \in \mathbb{R}^{64 \times 64}$ is the egocentric depth image with $58^\circ \times 58^\circ$ field of view. The action $a_t \in \mathbb{R}^{12}$ specifies the joints' target positions: ${q_d} = q_{\mathrm{stand}} + a_t$, where ${q_{\mathrm{stand}}}$ is the default joint positions when standing. The torques ${\tau}$ are calculated through a PD controller:
\begin{equation}
     \tau=K_{p}\left(q_{d}-q\right)+K_{d}\left(\dot{q}_{d}-\dot{q}\right),
\end{equation}
where $q$ and $\dot{q}$ are the joints' current positions and velocities, respectively. The target joint velocities $\dot{q}_{d}$ are set to 0 and $(K_p,K_d)$ are the parameters of the PD controller. 

\vspace{0.1cm}
\noindent \textbf{Reward Function.} The robot is trained to track a 3-dim command: $(v^{\mathrm{cmd}}_{x}, v^{\mathrm{cmd}}_{y}, \omega^{\mathrm{cmd}}_{z})$. To achieve this, we adopt a suite of reward functions similar to Cheng \etal~\cite{parkour2}. The main difference is that they manually select waypoints along a preset trajectory and compute the velocity-tracking reward based on the direction to the next waypoint. on the contrary, we use a simpler form of velocity-tracking reward to reduce human efforts in waypoint selection:
\begin{equation}
    r_{\mathrm{tracking}} = \mathrm{exp}\Big(\big(\mathrm{min}(v_{xy}, v^{\mathrm{cmd}}_{xy}+0.1) - v^{\mathrm{cmd}}_{xy}\big)^2/ \sigma \Big),
\end{equation}
where the clipping operation encourages the robot to follow the command most of the time, but it can also reach higher speeds when necessary, \eg jumping over gaps. We add additional penalties to avoid getting stuck or turning around the obstacles. Besides, we employ an AMP (Adversarial Motion Priors) \cite{amp} style reward to make the robot converge to a more natural behavior \cite{amp-wu}:
\begin{equation}
    r_{\mathrm{style}}(s, s') = \max [0, 1 - 0.25(D_{\psi}(s, s') - 1)^2],
    \label{eq:amp_reward}
\end{equation}
where $D_{\psi}$ is the discriminator trained to distinguish whether a state transition is from a reference dataset $\mathcal{D}_{\mathrm{ref}}$ or produced by the agent:
\begin{equation}
    \begin{split}
        \argmin_\psi \text{  } & \mathbb{E}_{(s, s') \sim \mathcal{D}_{\mathrm{ref}}}\left[(D_\psi(s, s') - 1)^2\right]\\
        +& \mathbb{E}_{(s, s') \sim \pi_\theta(s, a)}\left[(D_\psi(s, s') + 1)^2\right]\\
        +& {\frac{w^{\text{gp}}}{2}} \mathbb{E}_{(s, s') \sim \mathcal{D}_{\mathrm{ref}}}\left[\|\nabla_\psi D_\psi(s, s')\|^2\right].
    \end{split}
    \label{eq:disc_objective}
\end{equation}

\vspace{0pt}
\section{Experimental Results}
\label{sec:Experimental Result}

\begin{figure*}[htp]
\centering
        \vspace{5pt}
    \begin{minipage}[t]{0.31\textwidth}
        \centering
        \includegraphics[width=1\textwidth]{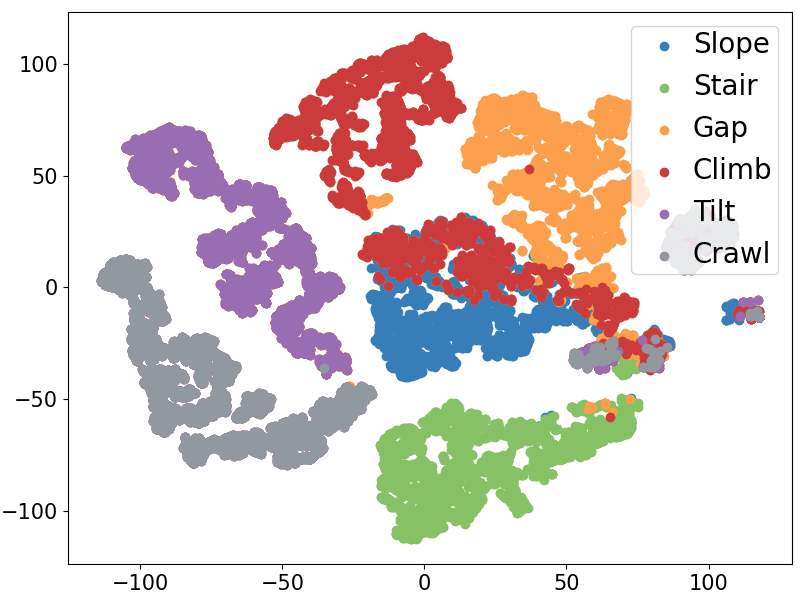}
        \vspace{-15pt}
        \caption{T-SNE result of recurrent state $h_t$ over six different terrains.}
        \label{fig:tsne}
        \end{minipage}
            \hspace{4pt}        
    \begin{minipage}[t]{0.66\textwidth}
        \includegraphics[width=1\textwidth]{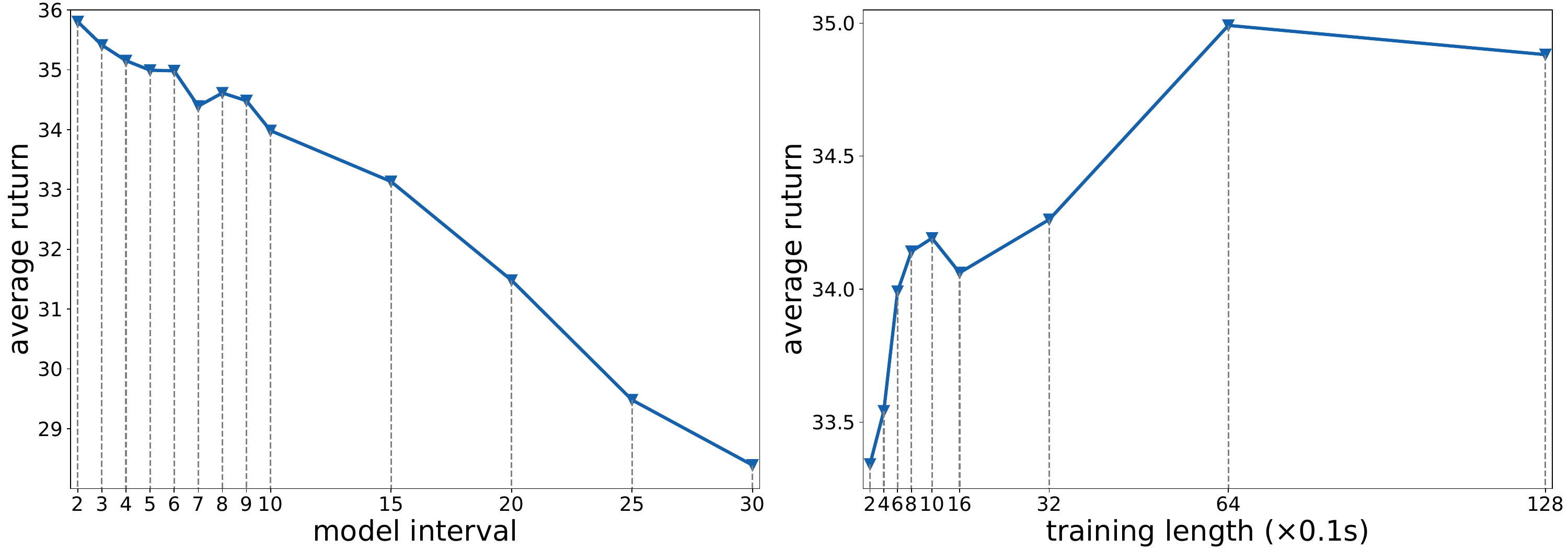}
        \vspace{-15pt}
        \caption{Average return over different world model interval $k$ (Left) and different lengths of training data (Right).}
        \label{fig:hyper}
    \end{minipage}

        \vspace{-15pt}
\end{figure*}

Our experiments aim to answer the following questions:
\begin{itemize}
	\item How does WMP perform compared with previous state-of-the-art methods in vision-based locomotion?
	\item Could a world model trained in a simulator predict real-world trajectory well?
	\item Could the achievement of WMP in simulation be well transferred to real robots?
\end{itemize}

To ensure a fair comparison, all the methods are trained using the same environment and reward functions described in Section~\ref{sec:Training Details}.

\subsection{Simulation Comparison}
\label{sec:Simulation Comparison}
To answer these questions, we first evaluate our method and baselines in terms of RL return and velocity tracking error over different terrains in simulation, where the velocity tracking error is defined as the mean square error between $v^{\mathrm{cmd}}_{xy}$ and $v_{xy}$. The baselines we compare to include:
\begin{itemize}
	\item \textbf{Teacher.} The teacher policy is trained with access to privileged information like scandots, serving as an oracle baseline.
	\item \textbf{Student.} We reproduce the student policy according to Cheng \etal~\cite{parkour2}, which utilizes a ConvNet-RNN to mimic the teacher's policy using depth images. 
	\item \textbf{Blind.} We ablate the depth image in the world model, resulting in a blind policy that gives actions purely based on proprioception.
	\item \textbf{WMP w/o Prop.} Similar to Blind, we remove the proprioception in the world model.
\end{itemize}

The comparison results are shown in Table~\ref{table:simulation comparison}. Note that the Teacher and Student baselines do not apply to Tilt and Crawl due to the limitation of scandots, as discussed in Section~\ref{sec:Introduction}. From the result, our method WMP achieves higher return and smaller velocity tracking error than baselines in most tasks. The performance gap between WMP and Teacher is much smaller than between Teacher and Student, revealing the superiority of WMP by leveraging the world model to extract proper information from past perceptions. Moreover, removing the depth image in the world model causes a severe performance dropping except for Slope, underscoring the importance of visual information for locomotion over challenging terrains \cite{egocentric, parkour1}. Besides, ablating the proprioception also decreases the performance slightly, which we attribute to the fact that predicting the proprioception can help capture the physical properties of the environment.

\subsection{Empirical Study}
\label{sec:Empirical Study}
This section provides empirical studies to understand the benefits of our method from the world model.

\vspace{0.1cm}
\noindent \textbf{Recurrent State Visualization.} We first collect the recurrent state $h_t$ over six terrains and visualize them in Figure~\ref{fig:tsne} to investigate whether $h_t$ contains enough information for versatile locomotion. As the t-SNE figure shows, the $h_t$ of different terrains holds clear boundaries, with only a slight overlap between Slope and Climb, since these two terrains have similar depth images and Climb can be considered as a $90^{\circ}$ Slope. From the visualization result, $h_t$ represents the terrains well and can help the policy take action according to the specific task.

\begin{figure}[htp]
	\centering
        \hspace{-2pt}
\includegraphics[width=0.47\textwidth]{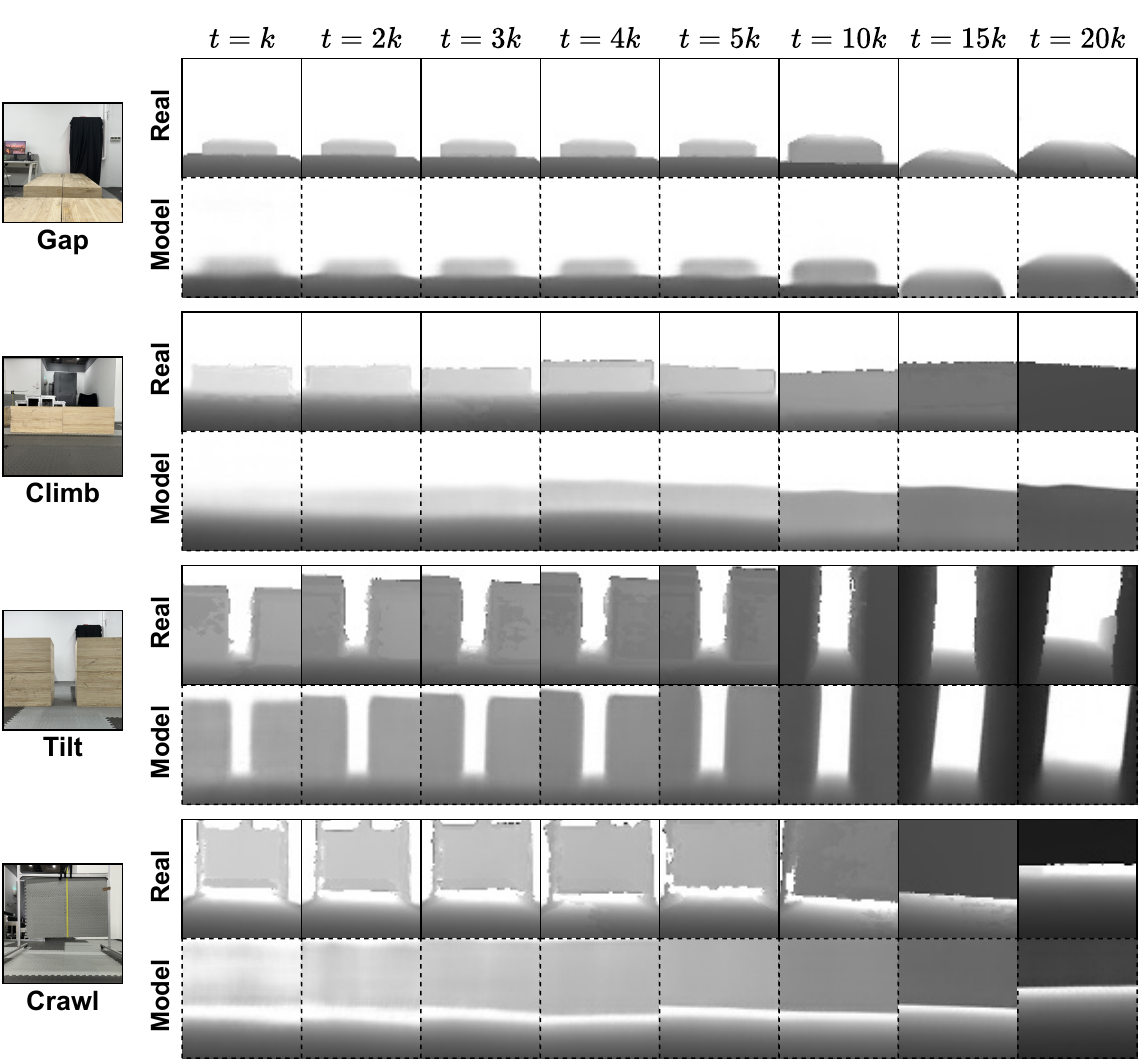}
	\vspace{-0pt}
	\caption{Real-world depth images and long-term predictions of depth images using the world model.}
	
	\vspace{-15pt}
	\label{fig:real world depth}
\end{figure}

\vspace{0.1cm}
\noindent \textbf{Model Interval.}  The model interval parameter $k$ affects both world model training and real-world deployment. To investigate its influence, we vary $k$ from 2 to 30. The results are shown in Figure~\ref{fig:hyper}. In general, a world model with a smaller interval obtains higher rewards in simulation because it enables the robot to respond to changes in its surroundings more quickly. However, unlike the ideal situation in the simulator, real-world applications have non-negligible latency for depth image acquiring and world model computing, taking around 40ms in total on A1 hardware. Therefore, we choose $k = 5$, \ie world model intervals of 0.1s, a trade-off between ideal performance and computational cost.

\vspace{0.1cm}
\noindent \textbf{Training Length.} During world model training, we randomly sample trajectory segments with fixed length $L$ and train the model to predict current perception based on previous ones in the segment. The length of the training data determines how long the historical information model can remember. In Figure~\ref{fig:hyper}, we conduct experiments with different training lengths. According to the result, training world models with 1-second segments is sufficient to achieve acceptable performance. This is consistent with our intuition: what the robot saw one second before is roughly under its feet. Further extending the horizon can help perceive the environment dynamics, but a segment that is too long may make it inefficient to back-propagate the gradient through RSSM. For this reason, we set the training length to 6.4 seconds in other experiments throughout the paper.

\begin{figure*}[htp]
	\centering
	\vspace{1pt}
\includegraphics[width=1.0\textwidth]{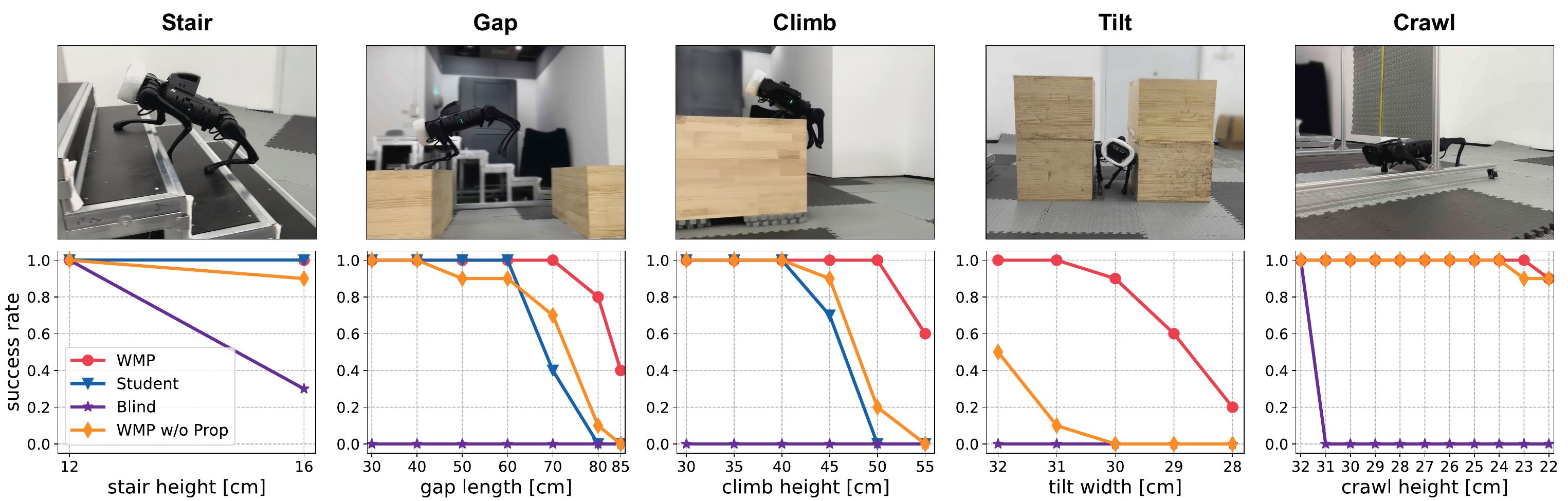}
	\vspace{-14pt}
	\caption{Real-world evaluation over multiple terrains with different difficulties. Success rates are calculated over ten trials.}
	
	\vspace{-15pt}
	\label{fig:real-world indoor}
\end{figure*}

\vspace{0.1cm}
\noindent \textbf{Real World Prediction.} While the world model performs well in simulation, whether the excellent performance can transfer to the real world remains to be justified. To verify this, we collect trajectories in the real world and use the model to predict the future, given the initial observation and action sequence without access to intermediate depth images. Results are shown in Figure~\ref{fig:real world depth}. From the result, a world model trained purely in the simulator can give accurate predictions for real-world trajectories, especially in the critical place it will pass. For example, in the Crawl task, the shape of the obstacle in the predicted depth image is different from the real ones since robots have never seen this shape of obstacles in the simulator. Nevertheless, the position and angle of the narrow crevice it can traverse are highly consistent in real and model images, highlighting the strength and generalization of latent space world modeling. This finding may help explain why our method exhibits smooth sim-to-real transfer.

\subsection{Real-world Evaluation}
\label{sec:Real-world Evaluation}
Subsequently, we apply WMP and other baselines to a real-world A1 robot. All methods are directly run on the onboard Jetson NX hardware without external computing devices. Depth images are read from the front Intel D435i camera at 60 Hz with a resolution of $424 \times 240$. We pre-process the noisy depth images with spatial and temporal filters to narrow the visual sim-to-real gap \cite{parkour1}. The processed images are then cropped and down-sampled to $64 \times 64$ and sent to the world model with 100ms latency. We set $K_p = 40, K_d = 1.0$ to make it consistent with the simulation setting. We select five terrains with different difficulties for a comprehensive evaluation, including Stair, Gap, Climb, Tilt, and Crawl, but exclude Slope, which is too easy to distinguish between methods. 

The success rates are listed in Figure~\ref{fig:real-world indoor}. From the comparison, the Student policy can traverse through the first three terrains with low difficulty but fails in more difficult cases, reflecting the performance gap between the Student policy and the optimal Teacher policy. In contrast, our method exhibits a more stable control behavior and successfully traverses more challenging terrains, including Tilt and Crawl, which the Student policy cannot tackle. To name a few, WMP can traverse Gap of 85cm (about 2.1x robot length), Climb of 55cm (about 2.2x robot height), and Crawl of 22cm (about 0.8x robot height), close to the hardest level in the simulator. This means that our method achieves smaller sim-to-real gaps through world modeling. Besides, ablating proprioception or images in the world model degrades performance to different degrees, demonstrating the advantage of physical and visual world modeling for locomotion. Please refer to the supplemental video for detailed comparisons.

\begin{figure}[htp]
	\centering
	\vspace{3pt}
        \hspace{-5pt}
\includegraphics[width=0.49\textwidth]{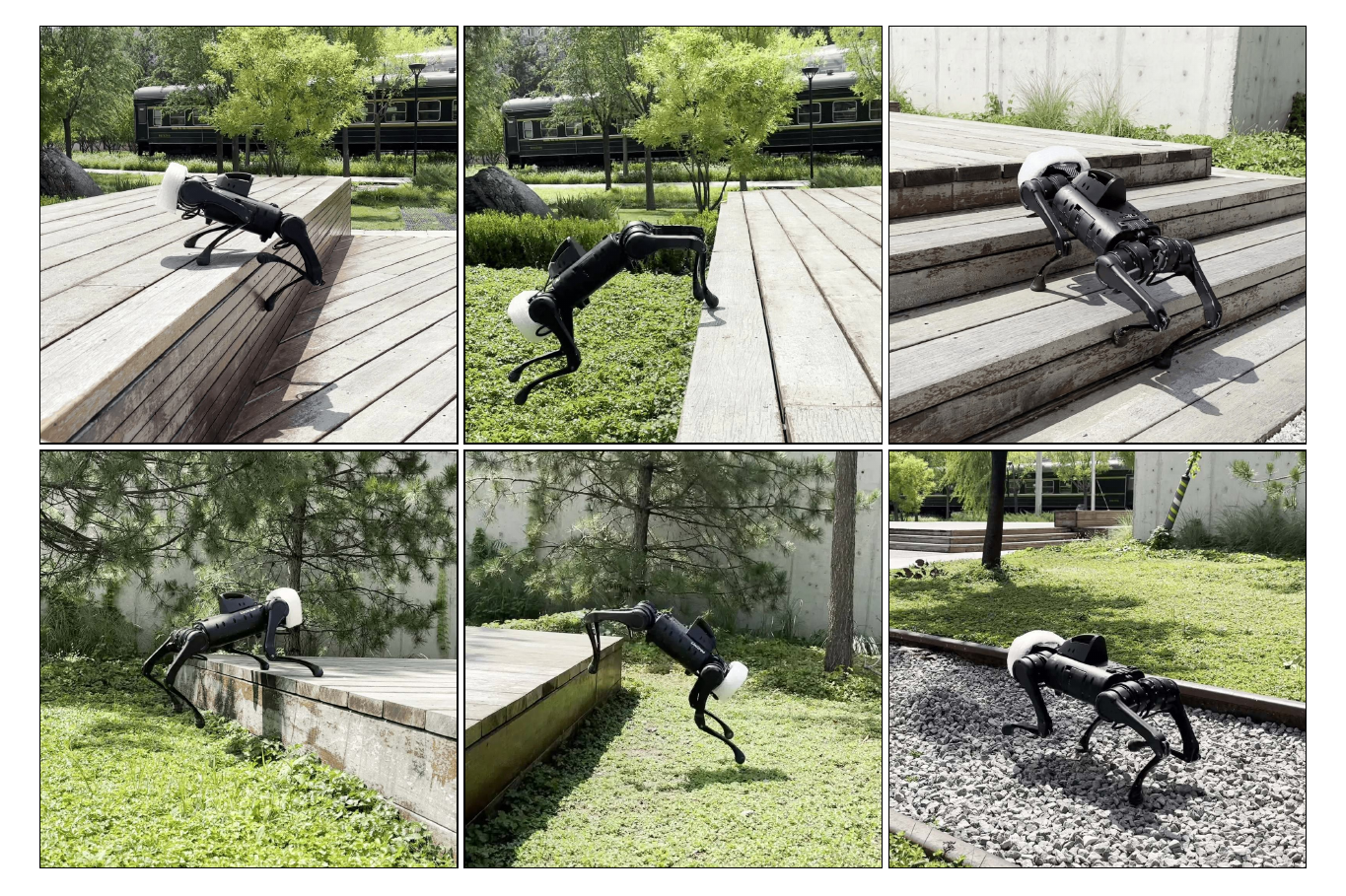}
	\vspace{-17pt}
	\caption{Snapshots of outdoor experiments.}
	
	\vspace{-15pt}
	\label{fig:real world outdoor}
\end{figure}

We also deploy our policy to outdoor environments in a park. Some snapshots are shown in Figure~\ref{fig:real world outdoor}. Our policy shows consistent behavior in outdoor and indoor environments and successfully goes up and down stairs, climbs platforms up to 45cm, and traverses grass and gravel, which further validates the generalization of our method.

\section{Conclusion}	
\label{sec:Conclusion}
In this paper, we present World Model-based Perception (WMP), a simple yet effective framework that combines MBRL with vision-based legged locomotion, drawing inspiration from the role of the mental model in animal cognition and decision-making. By leveraging the advanced world model, WMP outperforms previous state-of-the-art baselines in both simulation and real-world evaluation, achieving the best traversal performance on Unitree A1 robots. Further empirical analyses reveal that the main superiority of WMP lies in utilizing the world model to extract useful information from historical high-dimensional perceptions. We hope our method could provide insight into the emergence of a better natural learning paradigm for robots. For future work, it is tempting to train the world model with a mixture of simulated and real-world data, which holds the promise to construct a more realistic world model. Besides, it is also appealing to incorporate other forms of perception, like the sense of touch, into the world model to expand its applications.

\bibliographystyle{IEEEtran}
\bibliography{reference}
\end{document}